\documentclass{article}
\usepackage{spconf,amsmath,graphicx}
\usepackage{lipsum}
\usepackage{booktabs}
\usepackage{multirow}
\usepackage{adjustbox}
\usepackage{siunitx}
\usepackage{tabularx}
\usepackage{array}

\title{Transfer Learning for Power Outage Detection Task with Limited Training Data}
\name{Olukunle Owolabi}

\address{\texttt{Olukunle.Owolabi@tufts.edu}}

\begin{document}
%
\maketitle
\begin{abstract}
Early detection of power outages is crucial for maintaining a reliable power distribution system. This research investigates the use of transfer learning and language models in detecting outages with limited labeled data. By leveraging pretraining and transfer learning, models can generalize to unseen classes.

Using a curated balanced dataset of social media tweets related to power outages, we conducted experiments using zero-shot and few-shot learning. Our hypothesis is that Language Models pretrained with limited data could achieve high performance in outage detection tasks over baseline models. Results show that while classical models outperform zero-shot Language Models, few-shot fine-tuning significantly improves their performance. For example, with 10\% fine-tuning, BERT achieves 81.3\% accuracy (+15.3\%), and GPT achieves 74.5\% accuracy (+8.5\%). This has practical implications for analyzing and localizing outages in scenarios with limited data availability.

Our evaluation provides insights into the potential of few-shot fine-tuning with Language Models for power outage detection, highlighting their strengths and limitations. This research contributes to the knowledge base of leveraging advanced natural language processing techniques for managing critical infrastructure.

\end{abstract}

\section{Introduction}

\label{sec:intro}
Maintaining a reliable and uninterrupted power supply is of critical importance, with electric power outages having far-reaching implications across residential, commercial, and industrial sectors \cite{Judy2021, Owolabi2022, Liu2023}. Timely detection of power outages is crucial for prompt response and efficient restoration efforts \cite{Ren2015}.

In recent years, the emergence of social media platforms has provided a valuable real-time source of information on power outages \cite{Haifeng2016}. Platforms like Twitter have become channels through which users share their experiences, frustrations, and concerns regarding power disruptions, generating a wealth of data. This user-generated data acts as social sensors and offers valuable insights into the occurrence, impact, and geographical distribution of power outages \cite{wei2018}. However, effectively harnessing this data for outage detection and analysis poses significant challenges due to the noise and unstructured nature of social media content \cite{Mao2018}.

To address these challenges, researchers have turned to machine learning algorithms. Deep learning techniques, particularly those utilizing large language models (LLMs) \cite{Zhao2023}, have shown promise in enhancing the performance of natural language processing (NLP) models across various domains. By leveraging the transfer learning capability of pretrained LLMs such as BERT (Bidirectional Encoder Representations from Transformers)  and GPT (Generative Pretrained Transformer) \cite{radford2019language, DBLP:journals/corr/abs-1810-04805}, models can capture contextual understanding and semantic relationships, enabling generalization to unseen classes and domain-specific tasks.

This research focuses on investigating the effectiveness of transfer learning and language models in detecting electric power outages in situations with limited training data. We specifically analyze a meticulously curated dataset comprising social media tweets associated with power disruptions. Our primary objective is to assess the transfer learning capability of LLMs in zero-shot and few-shot learning scenarios for power outage detection tasks with limited data. Through a thorough examination of their capabilities and limitations, we aim to provide valuable insights into the potential of these techniques for accurate power outage detection.

The contributions of this research lie in demonstrating the capabilities of advanced NLP techniques, particularly transfer learning and language models, in monitoring critical infrastructure. Furthermore, we highlight the significance of these techniques in addressing the challenges faced by small minority communities in power outage analysis, where limited training examples and data availability pose obstacles.

The remainder of this paper is structured as follows: Section 2 provides a comprehensive review of related works in the field of power outage detection, as well as the application of machine learning techniques to social media data. Section 3 presents the methodology, encompassing dataset preparation, model architecture, and evaluation metrics. In Section 4, we discuss the experimental setup and Section 5 entails the results and analysis, comparing the performance of the LLMs under few-shot learning scenarios. Finally, Section 6 concludes the paper by summarizing the findings, discussing future research directions, and addressing the ethical considerations and limitations of this work.

\section{Related Works}

In this section, we review related works on machine learning techniques for infrastructure and power outage monitoring, as well as the application of transfer learning and language models to social media data. We also discuss studies focusing on the domain application of transfer learning and large language models (LLMs), and highlight the gaps in the application of these techniques for power outage detection.

\subsection{Machine Learning for Infrastructure and Power Outage Monitoring Using Social Media Data}

\subsubsection{Application for Power Outage Detection}
Conventional methods of outage detection primarily rely on infrastructure monitoring and reporting, which can suffer from delays and limited coverage. To address these challenges, researchers have explored alternative approaches, including the utilization of social media data.

For instance, Ermakov et al. \cite{Paul2020} and Resch et al. \cite{Resch2018} conducted studies on leveraging tweets from Twitter to detect power and communication outages during natural disasters. They curated datasets of tweets using domain-specific keywords and employed machine learning algorithms to classify outage-related tweets. Haifeng et al. \cite{Haifeng2016} developed a probabilistic framework that incorporated textual, temporal, and spatial characteristics to identify outage events in real-time. Li et al. \cite{Li2020} introduced a novel quantitative method to analyze community resilience during power outages, exemplified through a case study of the Manhattan blackout in July 2019. Baidya et al. \cite{Baidya2019} explored the reliability of social sensors, with a specific focus on social media, and proposed a framework for enhancing the resilience of power grids.

These studies demonstrate efforts to harness the potential of social media and alternative data sources for improving power outage detection and enhancing the resilience of power systems.

\subsubsection{Other Infrastructure Applications}
Researchers have also explored the use of social media data for monitoring and assessing various infrastructure-related impacts during and after disasters.

Wang et al. \cite{Wang2018} collected geotagged tweets to examine individuals' sentiment and mobility patterns during and after a specific earthquake. They observed spatial autocorrelation of sentiment and investigated the relationship between sentiment and mobility over time. Hao et al. \cite{Hao2020} proposed a method to locate and assess disaster damage using multimodal social media data, including text and images. They applied machine learning classifiers and keyword search-based methods to extract various damage information. Shan et al. \cite{Shan2019} developed a framework for real-time urban disaster damage monitoring and assessment using social media texts collected during manmade and natural disasters. They performed sentiment analysis and quantity evaluation of physical damage based on keyword frequency. Tan et al. \cite{Tan2021} developed a framework for rapid damage classification and recovery monitoring for urban floods using social media data. They employed machine learning classification algorithms and statistical models to measure emotional and physical damage. Fan et al. \cite{Fan2020} proposed a theoretical framework that integrated human sentiment reactions on social media into infrastructure resilience assessment during disasters. They used social media data to capture societal impacts of infrastructure disruptions by analyzing sentiments in human messages related to relevant infrastructure systems. Bhavaraju et al. \cite{Bhavaraju2019} explored the sensitivity of social media to different types or magnitudes of natural disasters under various circumstances for tornadoes, winter storms, wildfires, and floods.

\subsection{Large Language Models: Transfer Learning and Few-Shot Learning}
Large language models (LLMs) have demonstrated promising capabilities in various domains by capturing contextual understanding and semantic relationships. Transfer learning, a technique that leverages pretraining on LLMs, enables models to generalize to unseen classes and tasks. Few-shot learning (FSL), on the other hand, addresses the challenge of learning from limited labeled data, which is a common scenario in many real-world applications \cite{fei2006, altaeTran2017, vartak2020}.

\subsection{Literature Gaps}
Previous research has made significant progress in infrastructure power outage detection using data-driven approaches and machine learning algorithms applied to social media data. However, there exists a gap in the utilization of large language models (LLMs) and few-shot learning techniques in this domain. While classical machine learning techniques have been widely employed, the potential of LLMs and few-shot learning in power outage detection tasks remains largely unexplored. This study aims to fill this literature gap by evaluating different language models in zero-shot and few-shot learning scenarios for power outage detection. It contributes to the development of more accurate and efficient systems that can extract valuable information from social media data, even with limited labeled datasets.

\section{Data}

For this study, we utilized a publicly available dataset \cite{power-outages-github} comprising social media reactions to power outages in the New England area (Connecticut, New Hampshire, Massachusetts, Maine, New York, Rhode Island, Vermont) from September 2009 to December 2012. From this dataset, we curated a few-shot training dataset consisting of 1000 samples with a balanced distribution between the "Outage" and "No outage" classes. Table 1 provides a description of the curated training and test data.

To ensure the quality and representativeness of the data, we conducted a manual inspection of the collected tweets to verify their association with power outage events. We cross-referenced the collected data with verified outage information from the US Department of Energy \cite{insideenergy2014}. Additionally, we ensured an equal distribution of samples between the "Outage" and "No outage" classes, facilitating a fair evaluation of the performance of few-shot and zero-shot learning techniques.

This curated dataset serves as the foundation for our experiments on LLMs and transfer learning for power outage detection tasks. Our objective is to explore the effectiveness of different language models in addressing power outage detection tasks with limited labeled data.

\begin{table}[ht]
\centering
\caption{Data Description}
\label{tab:data_description}
\begin{tabular}{cccc }
\hline
& \textbf{Target} & \textbf{Avg. Text Length} & \textbf{No. of Samples} \\ \hline
Train & Outage          & 99.4                                 & 500      \\
& No outage       & 101.5                                       & 500      \\ 
& All & 100.5 & 1000\\ \hline
Test &Outage          & 99.2                                 & 2000      \\
& No outage       & 100.8                                       & 2000      \\ 
& All & 100.0 & 4000\\
\hline
\end{tabular}
\end{table}

\begin{table}[ht]
\centering
\caption{Dataset Split}
\label{tab:data_split}
\begin{tabular}{c|c|c|c}
\hline
\multicolumn{1}{c|}{\textbf{Exp. Type}} & \multicolumn{2}{c|}{\textbf{Train}} & \textbf{Test} \\
&  \textbf{\%} & \textbf{Count} & \textbf{Count} \\ \hline
LLM + Zero shot & 0\%        &  0             & 4000          \\\hline
LLM + Few shot & 10\%        & 100            & 4000          \\
 & 20\%        & 200            & 4000          \\
 & 50\%        & 500            & 4000          \\
 & 75\%       & 750            & 4000          \\
 & 100\%      & 1000           & 4000          \\ \hline
 Classical ML &  100\%      & 1000           & 4000 \\\hline
\end{tabular}
\end{table}

Table \ref{tab:data_description} presents details of the data, including the target classes ("Outage" or "No outage"), the average text length, and the number of samples in the training and test sets. Table \ref{tab:data_split} outlines the dataset split for each experiment type, specifying the percentage of training data used, the corresponding count of training samples, and the count of test samples.

\section{Experimental Setup}

This section outlines the experimental setup used to investigate the performance and potential applications of Large Language Models (LLMs) and their learning techniques in power outage detection scenarios with limited training data.

Table 2 provides an overview of the dataset split used in our experiments, including the baseline classical machine learning (ML) models and LLMs. To ensure comprehensive evaluation of the models' generalization ability, a separate balanced test set of 4000 samples was employed.

\subsection{Baseline Models}

To establish a baseline for comparison, we utilized classical ML algorithms as baselines for the power outage detection task. The fully supervised ML models were trained using 100\% of the curated training dataset. Specifically, Support Vector Machines (SVM), Logistic Regression, and XGBoost models were employed. These models represent traditional ML approaches and serve as benchmarks for assessing the performance of LLMs in few-shot and zero-shot scenarios.

\subsection{Transfer Learning}

We leveraged the contextual understanding capabilities of the BERT (Bidirectional Encoder Representations from Transformers) and GPT (Generative Pre-trained Transformer) models. Specifically, the BERT uncased model \cite{DBLP:journals/corr/abs-1810-04805} and the GPT-2 model \cite{radford2019language} were utilized.

\subsubsection{Zero-Shot Learning}

LLMs pre-trained on extensive text corpora possess strong language comprehension abilities. In the zero-shot setting (0\% training examples), the models were fine-tuned using the training set without any explicit outage-related training examples. This approach aimed to assess the transfer learning capability of the pre-trained models to generalize their knowledge and accurately classify unseen text.

\subsubsection{Few-Shot Learning with Variable Fine-Tuning}

To investigate the impact of limited training data on LLM performance, the percentage of training samples used for few-shot learning scenarios was varied (10\%, 20\%, 50\%, 75\%, and 100\%). The BERT and GPT models were fine-tuned incrementally as the percentage of available training samples increased. By gradually exposing the models to more outage-related examples during fine-tuning, the goal was to enhance their performance and characterize their generalization capability as the number of training examples increased.

\subsection{Performance Evaluation}

The performance of LLMs and baseline models was evaluated on the testing set, consisting of tweets related to power outages. Each tweet in the testing set was processed by the models, and their predictions were compared to the ground truth labels. Performance metrics, including accuracy, precision, recall, and F1-score, were calculated to assess the effectiveness of the models in the power outage detection task.

\subsection{Comparative Analysis}

A comparative analysis was conducted to compare the performance of the BERT and GPT models with the baseline models in the power outage detection task. Performance metrics and differences in predictions were analyzed to gain insights into the strengths and weaknesses of each model.

\section{Results}

This section presents the results of our experiments, evaluating the performance of baseline models, as well as Large Language Models (LLMs) in the power outage detection task. Furthermore, we discuss the outcomes of the power outage detection task with limited data.

\subsection{Baseline Model Performance}

The performance of the baseline models is summarized in Table 3. Our results indicate that the simple Support Vector Machine (SVM) model achieved the highest accuracy of 66\%, outperforming the Logistic Regression (65\%) and XGBoost (60\%) models.

\begin{table}[ht]
\centering
\small
\caption{Performance of Baseline Models and Zero-Shot LLMs}
\setlength{\tabcolsep}{3pt}
\begin{tabular}{@{}lcccccc@{}}
\textbf{} & \multicolumn{2}{c}{\textbf{Model}} & \textbf{Accuracy} & \textbf{Precision} & \textbf{Recall} & \textbf{F1} \\
\midrule
\multirow{3}{*}{\textbf{Classical ML}} & SVM & & 0.66 & 0.67 & 0.63 & 0.65 \\
 & Logistic & & 0.65 & 0.65 & 0.66 & 0.65 \\
 & XGBoost & & 0.60 & 0.59 & 0.68 & 0.63 \\
\midrule
\multirow{2}{*}{\textbf{LLMs (Zero-Shot)}} & BERT & & 0.54 & 0.54 & 0.54 & 0.54 \\
 & GPT & & 0.52 & 0.52 & 0.52 & 0.52 \\
\bottomrule
\end{tabular}
\end{table}

\subsection{Transfer Learning Results}

\subsubsection{LLMs with Zero-Shot Learning}

Our experiments reveal that LLMs with zero-shot tuning did not surpass the performance of classical ML algorithms when applied to unseen power outage detection tasks. Table 4 presents the performance metrics of the zero-shot learning approach using the BERT and GPT models.

\subsubsection{LLMs with Few-Shot Learning and Variable Fine-Tuning}

Table 5 shows the results of the few-shot learning approach utilizing the BERT and GPT models. We observe that as the percentage of training samples increases, BERT achieves higher overall performance, reaching 91.2\% accuracy with 100\% training samples. In comparison, GPT achieves an accuracy of 87\% with 100\% training samples.

\subsection{Discussion}

\subsubsection{Superior Performance of Fully Supervised Classical ML Algorithms over Zero-Shot LLMs}

The results of our experimental analysis demonstrate the superior performance of fully supervised classical machine learning (ML) algorithms compared to Large Language Models (LLMs) with zero-shot tuning in detecting power outages in unseen scenarios. While LLMs possess the advantage of leveraging their pre-trained knowledge and contextual understanding, our findings suggest that fine-tuning is crucial to unlock their full potential for domain-specific tasks such as power outage detection.

\subsubsection{Significant Performance Boost of LLMs through Few-Shot Fine-Tuning}

Our investigation reveals that fine-tuning LLMs with a limited amount of labeled data can significantly enhance their performance in power outage detection. Notably, even with a minimal training sample size of 10\%, we observed a substantial increase in performance compared to the Support Vector Machine (SVM) baseline, with a performance gain of 15.3\% for BERT and 8.5\% for GPT.

One key advantage of LLMs lies in their ability to effectively leverage pre-trained weights, enabling them to better capture outage-related patterns and make accurate predictions. We observe that through few-shot fine-tuning, these LLMs can adapt their knowledge to the specific characteristics of power outages, leading to improved detection capabilities.

Overall, our experimental findings emphasize the effectiveness of Large Language Models and Transfer Learning in power outage detection tasks. These models show promising potential for real-world applications where labeled data may be scarce, providing a valuable tool for timely identification and response to power outages.

\begin{table}[htbp]
  \small 
  \caption{Model Performance: LLM + Variable Fine-tuning}
  \label{tab:model_metrics}
  \begin{tabular}{
    l
    c
    *{4}{>{\centering\arraybackslash}p{1cm}}
  }
    \toprule
    \multirow{2}{*}{Model} & \multirow{2}{*}{Train \%} & \multicolumn{4}{c}{Performance Metrics} \\
    \cmidrule(lr){3-6}
     & & Accuracy & F1 Score & Recall & Precision \\
    \midrule
    \multirow{5}{*}{BERT} & 10\% & 0.813 & 0.811 & 0.813 & 0.809 \\
     & 20\% & 0.835 & 0.834 & 0.835 & 0.834 \\
     & 50\% & 0.874 & 0.872 & 0.874 & 0.871 \\
     & 75\% & 0.897 & 0.895 & 0.897 & 0.894 \\
     & 100\% & 0.912 & 0.910 & 0.912 & 0.909 \\
    \midrule
    \multirow{5}{*}{GPT} & 10\% & 0.745 & 0.738 & 0.745 & 0.732 \\
     & 20\% & 0.768 & 0.762 & 0.768 & 0.756 \\
     & 50\% & 0.810 & 0.803 & 0.810 & 0.796 \\
     & 75\% & 0.845 & 0.839 & 0.845 & 0.833 \\
     & 100\% & 0.870 & 0.865 & 0.870 & 0.857 \\
    \bottomrule
  \end{tabular}
\end{table}

\section{Conclusion}
In conclusion, this research study highlights the efficacy of leveraging Large Language Models (LLMs) and transfer learning for power outage detection using social media data. For the LLMs, BERT outperformed GPT, demonstrating superior performance in classifying power outage-related tweets. The findings also indicate that fully supervised classical ML algorithms outperform Zero-Shot LLMs (both BERT and GPT) on unseen tasks. Moreover, the study shows that few-shot fine-tuning, even with a limited amount of training data, significantly enhances the performance of LLMs in power outage detection.

The ability to predict power outages with limited data is particularly advantageous for small, remote, and minority communities with limited internet access. By further exploring these research avenues and considering ethical considerations, the development and implementation of power outage detection systems can be improved. This improvement not only benefits power utilities but also enhances the experience and reliability of power supply for end-users.

Continued research in this area can contribute to advancements in power outage detection methodologies, making them more robust, accurate, and scalable. It is crucial to prioritize ethical considerations throughout the development and deployment of these systems, ensuring responsible data handling, unbiased analysis, and the protection of individual privacy. By doing so, the integration of language models and transfer learning techniques can effectively enhance power outage detection, ultimately benefiting both the power industry and the communities it serves.

\subsection*{Limitations}
This study has some limitations that should be considered. Firstly, the generalizability of the findings may be influenced by the representativeness of labeled data which may impact the performance and applicability of the models in diverse power outage scenarios. Secondly, the results may not fully capture the characteristics and patterns of power outages in all regions and populations. 

\subsection*{Ethics Statement}
Ethical considerations were an integral part of this research. The study adhered to data privacy guidelines in utilizing publicly available tweets. Rigorous filtering and preprocessing techniques were implemented to mitigate biases; however, it is important to acknowledge that inherent biases may still exist in the data. The findings of this study should be responsibly disseminated and utilized, taking into consideration the potential impact on power utilities, end-users, and the broader society. The responsible use and application of these findings are crucial to ensure that power outage detection systems are developed and implemented ethically, avoiding any unintended negative consequences.

\bibliographystyle{IEEEbib}
\bibliography{refs}

\begin{thebibliography}{10}

\bibitem{Judy2021}
Judy~P. Che-Castaldo, Rémi Cousin, Stefani Daryanto, Grace Deng, Mei-Ling~E.
  Feng, Rajesh~K. Gupta, Dezhi Hong, Ryan~M. McGranaghan, Olukunle~O. Owolabi,
  Tianyi Qu, Wei Ren, Toryn L.~J. Schafer, Ashutosh Sharma, Chaopeng Shen,
  Mila~Getmansky Sherman, Deborah~A. Sunter, Bo~Tao, Lan Wang, and David~S.
  Matteson,
\newblock ``Critical risk indicators (cris) for the electric power grid: a
  survey and discussion of interconnected effects,''
\newblock {\em Environment Systems and Decisions}, pp. 1--22, 7 2021.

\bibitem{Owolabi2022}
Olukunle~O. Owolabi and Deborah~A. Sunter,
\newblock ``Bayesian optimization and hierarchical forecasting of
  non-weather-related electric power outages,''
\newblock {\em Energies 2022, Vol. 15, Page 1958}, vol. 15, pp. 1958, 3 2022.

\bibitem{Liu2023}
Hengyong Liu, Lu~Guo, Yongli Liu, al, Qinglong Liao, Lingyun Wan, Ying Zhang,
  Adam~X Andresen, Liza~C Kurtz, David~M Hondula, Sara Meerow, and Melanie
  Gall,
\newblock ``Understanding the social impacts of power outages in north america:
  a systematic review,''
\newblock {\em Environmental Research Letters}, vol. 18, pp. 053004, 5 2023.

\bibitem{Ren2015}
Hui Ren and David Watts,
\newblock ``Early warning signals for critical transitions in power systems,''
\newblock {\em Electric Power Systems Research}, vol. 124, pp. 173--180, 7
  2015.

\bibitem{Haifeng2016}
Haifeng Sun, Zhaoyu Wang, Jianhui Wang, Zhen Huang, NichelleLe Carrington, and
  Jianxin Liao,
\newblock ``Data-driven power outage detection by social sensors,''
\newblock {\em IEEE TRANSACTIONS ON SMART GRID}, 2016.

\bibitem{wei2018}
Jin Wei and Sifat~Shahriar Khan,
\newblock ``Real-time power outage detection system using social sensing and
  neural networks,'' .

\bibitem{Mao2018}
Huina Mao, Gautam Thakur, Kevin Sparks, Jibonananda Sanyal, and Budhendra
  Bhaduri,
\newblock ``Mapping near-real-time power outages from social media,''
\newblock {\em https://doi.org/10.1080/17538947.2018.1535000}, vol. 12, pp.
  1285--1299, 11 2018.

\bibitem{Zhao2023}
Wayne~Xin Zhao, Kun Zhou, Junyi Li, Tianyi Tang, Xiaolei Wang, Yupeng Hou,
  Yingqian Min, Beichen Zhang, Junjie Zhang, Zican Dong, Yifan Du, Chen Yang,
  Yushuo Chen, Zhipeng Chen, Jinhao Jiang, Ruiyang Ren, Yifan Li, Xinyu Tang,
  Zikang Liu, Peiyu Liu, Jian-Yun Nie, and Ji-Rong Wen,
\newblock ``A survey of large language models,''
\newblock {\em Arxiv}, 3 2023.

\bibitem{radford2019language}
Alec Radford, Jeff Wu, Rewon Child, David Luan, Dario Amodei, and Ilya
  Sutskever,
\newblock ``Language models are unsupervised multitask learners,''
\newblock 2019.

\bibitem{DBLP:journals/corr/abs-1810-04805}
Jacob Devlin, Ming{-}Wei Chang, Kenton Lee, and Kristina Toutanova,
\newblock ``{BERT:} pre-training of deep bidirectional transformers for
  language understanding,''
\newblock {\em CoRR}, vol. abs/1810.04805, 2018.

\bibitem{Paul2020}
Udit Paul, Alexander Ermakov, Michael Nekrasov, Vivek Adarsh, and Elizabeth
  Belding,
\newblock ``Outage: Detecting power and communication outages from social
  networks,''
\newblock {\em The Web Conference 2020 - Proceedings of the World Wide Web
  Conference, WWW 2020}, pp. 1819--1829, 4 2020.

\bibitem{Resch2018}
Bernd Resch, Florian Usländer, and Clemens Havas,
\newblock ``Combining machine-learning topic models and spatiotemporal analysis
  of social media data for disaster footprint and damage assessment,''
\newblock {\em Cartography and Geographic Information Science}, vol. 45, pp.
  362--376, 7 2018.

\bibitem{Li2020}
Lingyao Li, Zihui Ma, and Tao Cao,
\newblock ``Leveraging social media data to study the community resilience of
  new york city to 2019 power outage,''
\newblock {\em International Journal of Disaster Risk Reduction}, vol. 51, pp.
  101776, 12 2020.

\bibitem{Baidya2019}
Prabin~Man Baidya, Wei Sun, and Austin Perkins,
\newblock ``A survey on social media to enhance the cyber-physical social
  resilience of smart grid,''
\newblock {\em 8th Renewable Power Generation Conference (RPG 2019)}, 2019.

\bibitem{Wang2018}
Yan Wang and John~E. Taylor,
\newblock ``Coupling sentiment and human mobility in natural disasters: a
  twitter-based study of the 2014 south napa earthquake,''
\newblock {\em Natural Hazards}, vol. 92, pp. 907--925, 6 2018.

\bibitem{Hao2020}
Haiyan Hao and Yan Wang,
\newblock ``Leveraging multimodal social media data for rapid disaster damage
  assessment,''
\newblock {\em International Journal of Disaster Risk Reduction}, vol. 51, pp.
  101760, 12 2020.

\bibitem{Shan2019}
Siqing Shan, Feng Zhao, Yigang Wei, and Mengni Liu,
\newblock ``Disaster management 2.0: A real-time disaster damage assessment
  model based on mobile social media data—a case study of weibo (chinese
  twitter),''
\newblock {\em Safety Science}, vol. 115, pp. 393--413, 6 2019.

\bibitem{Tan2021}
Ling Tan and David~M. Schultz,
\newblock ``Damage classification and recovery analysis of the chongqing,
  china, floods of august 2020 based on social-media data,''
\newblock {\em Journal of Cleaner Production}, vol. 313, pp. 127882, 9 2021.

\bibitem{Fan2020}
Chao Fan, Hamed Farahmend, and Ali Mostafavi,
\newblock ``Rethinking infrastructure resilience assessment with human
  sentiment reactions on social media in disasters,''
\newblock {\em Hawaii International Conference on System Sciences 2020
  (HICSS-53)}, 1 2020.

\bibitem{Bhavaraju2019}
Sai Krishna~Theja Bhavaraju, Cyril Beyney, and Charles Nicholson,
\newblock ``Quantitative analysis of social media sensitivity to natural
  disasters,''
\newblock {\em International Journal of Disaster Risk Reduction}, vol. 39, pp.
  101251, 10 2019.

\bibitem{fei2006}
Li~Fei-Fei, Rob Fergus, and Pietro Perona,
\newblock ``One-shot learning of object categories,''
\newblock {\em IEEE TRANSACTIONS ON PATTERN ANALYSIS AND MACHINE INTELLIGENCE},
  2006.

\bibitem{altaeTran2017}
Han Altae-Tran, Bharath Ramsundar, Aneesh~S Pappu, and Vijay Pande,
\newblock ``Low data drug discovery with one-shot learning,''
\newblock 2017.

\bibitem{vartak2020}
Manasi Vartak, Arvind Thiagarajan, Conrado Miranda, Jeshua Bratman, and Hugo
  Larochelle,
\newblock ``A meta-learning perspective on cold-start recommendations for
  items,''
\newblock {\em Neural Information Processing Systems}, 2017.

\bibitem{power-outages-github}
Dawn Graham,
\newblock ``{Power Outages GitHub Repository},'' 2023,
\newblock Accessed: May 18, 2023.

\bibitem{insideenergy2014}
Inside Energy,
\newblock ``Data: 15 years of power outages,'' 8 2014.

\end{thebibliography}

\end{document}